\documentclass[letterpaper,10pt,conference]{ieeeconf}
\IEEEoverridecommandlockouts
\usepackage{iros}
\usepackage{bm}

\acrodef{slam}[SLAM]{simultaneous localization and mapping}
\acrodef{tamp}[TAMP]{task and motion planning}
\acrodef{urdf}[URDF]{unified robot description format}
\acrodef{pt}[\emph{pt}]{parse tree}
\acrodef{iou}[IoU]{intersection over union}
\acrodef{map}[mAP]{mean average precision}

\let\oldnl\nl
\newcommand{\nosemic}{\SetEndCharOfAlgoLine{\relax}}
\newcommand{\nonl}{\renewcommand{\nl}{\let\nl\oldnl}}
\newcolumntype{x}{>{\columncolor{MistyRose}}c}
\newcolumntype{y}{>{\columncolor{LightCyan1}}c}

\title{\LARGE \bf Reconstruction of Part-level Interactive Scenes using Primitive Shapes}%
\title{\LARGE \bf Part-level Scene Reconstruction Affords Robot Interaction}%

\author{Zeyu Zhang$^{1,2\,*}$, Lexing Zhang$^{1,3\,*}$, Zaijin Wang$^{1}$, Ziyuan Jiao$^{1,2}$, Muzhi Han$^{1,2}$,\vspace{3pt}\\Yixin Zhu$^{4}$, Song-Chun Zhu$^{1,3,4}$, Hangxin Liu$^{1\,\dagger}$
\thanks{$^{*}$ Z. Zhang and L. Zhang contributed equally to this work. Emails: \tt{zeyuzhang@ucla.edu}, \tt{zhanglexing@bigai.ai}.}
\thanks{$^\dagger$ Corresponding author. Email: \tt{liuhx@bigai.ai}.}%
\thanks{%
    $^{1}$ National Key Laboratory of General Artificial Intelligence, Beijing Institute for General Artificial Intelligence (BIGAI).
    $^{2}$ Center for Vision, Cognition, Learning, and Autonomy (VCLA), Statistics Department, UCLA.
    $^{3}$ School of Intelligence Science and Technology, Peking University.
    $^{4}$ Institute for Artificial Intelligence, Peking University.}
}%
\begin{document}

\maketitle
\thispagestyle{empty}
\pagestyle{empty}

\begin{abstract}
Existing methods for reconstructing interactive scenes primarily focus on replacing reconstructed objects with CAD models retrieved from a limited database, resulting in significant discrepancies between the reconstructed and observed scenes. To address this issue, our work introduces a part-level reconstruction approach that reassembles objects using primitive shapes. This enables us to precisely replicate the observed physical scenes and simulate robot interactions with both rigid and articulated objects. By segmenting reconstructed objects into semantic parts and aligning primitive shapes to these parts, we assemble them as CAD models while estimating kinematic relations, including parent-child contact relations, joint types, and parameters. Specifically, we derive the optimal primitive alignment by solving a series of optimization problems, and estimate kinematic relations based on part semantics and geometry. Our experiments demonstrate that part-level scene reconstruction outperforms object-level reconstruction by accurately capturing finer details and improving precision. These reconstructed part-level interactive scenes provide valuable kinematic information for various robotic applications; we showcase the feasibility of certifying mobile manipulation planning in these interactive scenes before executing tasks in the physical world.
\end{abstract}

\section{Introduction}

Reconstructing surroundings is critical for robots, enabling them to understand and interact with their environments. However, traditional scene reconstruction methods primarily focus on generating static scenes, represented by sparse landmarks~\cite{pronobis2012large,yang2019cubeslam}, occupancy grids~\cite{dissanayake2001solution}, surfels~\cite{mccormac2017semanticfusion,hoang2020panoptic}, volumetric voxels~\cite{grinvald2019volumetric,mccormac2018fusion++}, or semantic objects~\cite{yang2019cubeslam}. These representations lack the ability to capture the dynamic nature of robot operations and limit the complexity of tasks that can be performed, such as interactions with objects beyond pick-and-place. This limitation calls for a new approach that places interactions at the core of scene reconstruction.

\begin{figure}[t!]
    \centering
    \begin{subfigure}[b]{\linewidth}
        \centering
        \includegraphics[width=\linewidth]{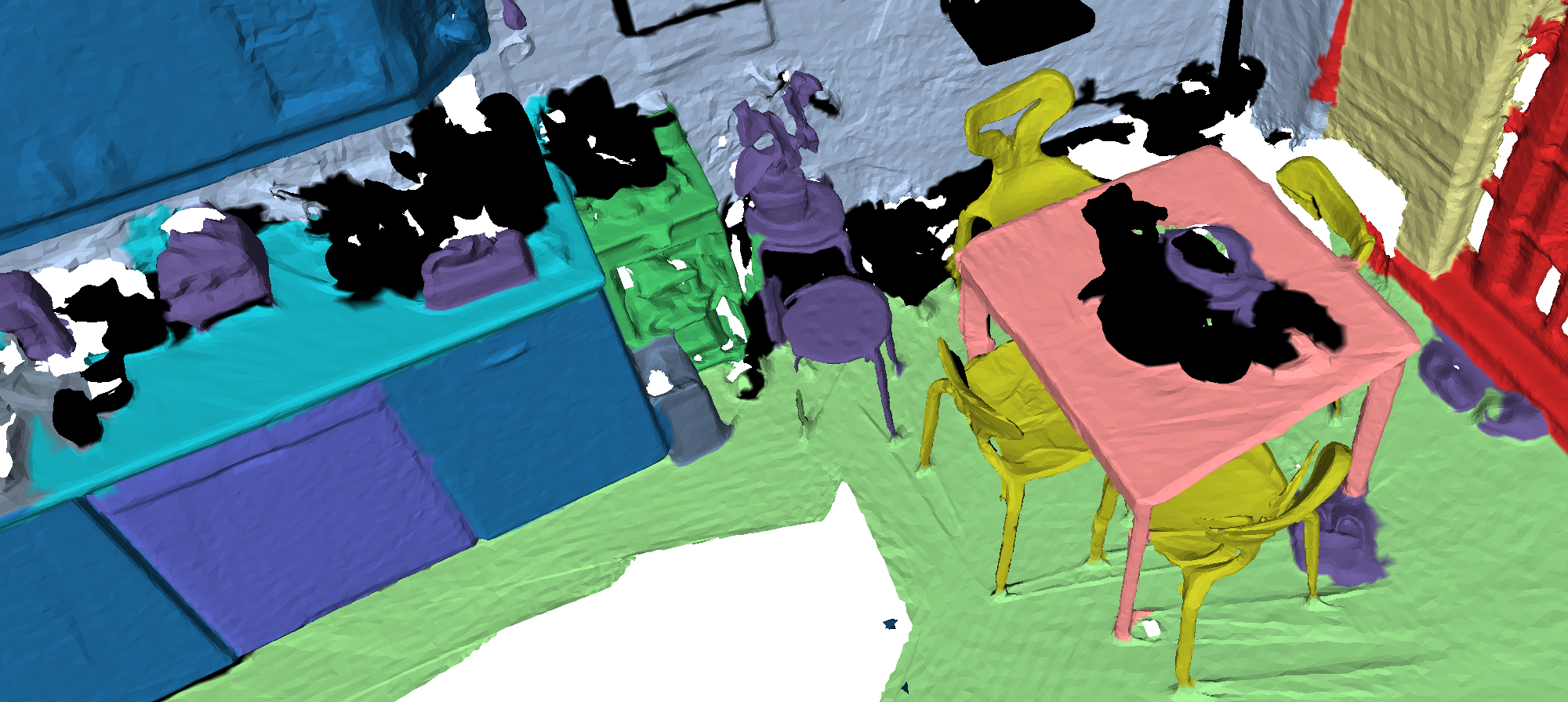}
        \caption{Panoptic mapping}
        \label{fig:motive_panoptic}
    \end{subfigure}
    \\
    \begin{subfigure}[b]{\linewidth}
        \centering
        \includegraphics[width=\linewidth]{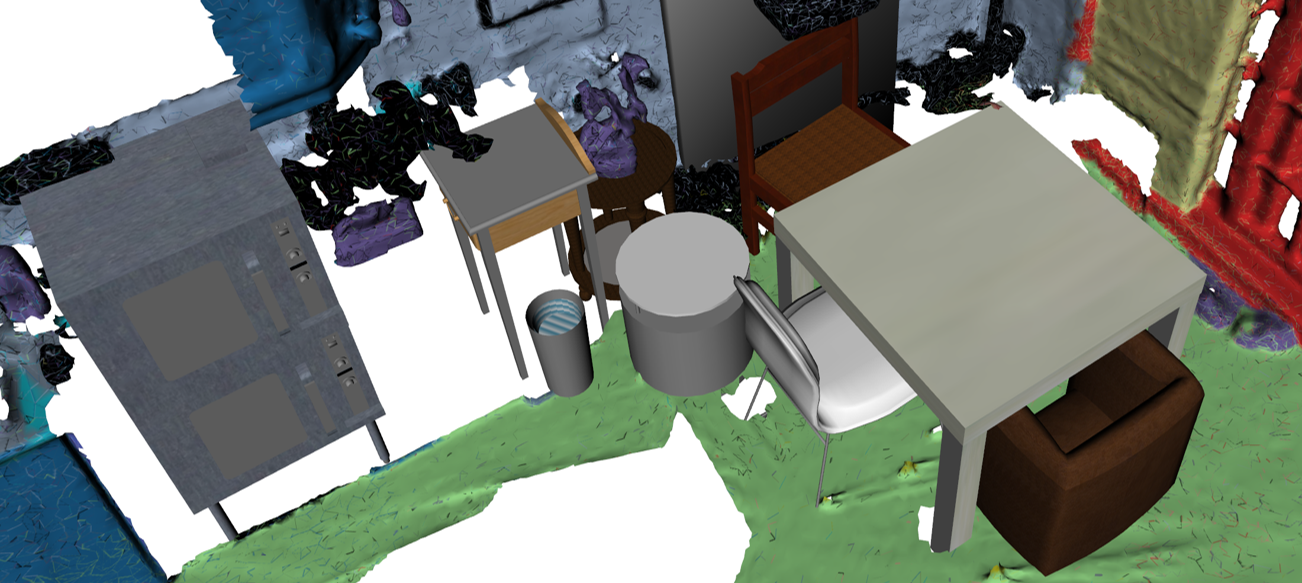}
        \caption{Scene reconstruction with object-level CAD replacement~\cite{han2021reconstructing}}
        \label{fig:motive_object}
    \end{subfigure}
    \\
    \begin{subfigure}[b]{\linewidth}
        \centering
        \includegraphics[width=\linewidth]{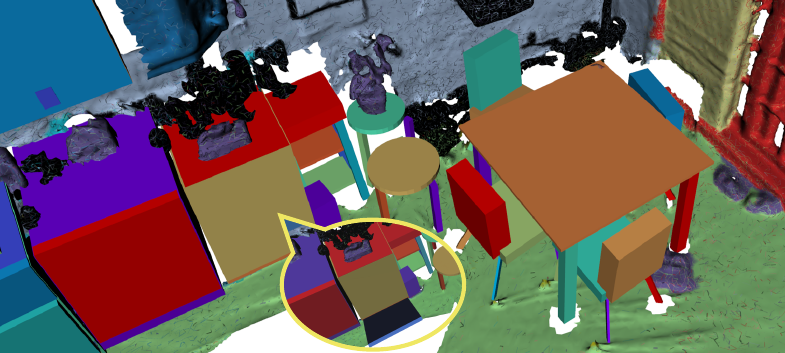}
        \caption{Scene reconstruction with part-level CAD replacement (ours)}
        \label{fig:motive_part}
    \end{subfigure}
    \caption{\textbf{Reconstructing interactive scenes.} (a) The initial step involves generating a panoptic mapping result with recognized and segmented object instances. (b) Existing methods~\cite{han2021reconstructing,han2022scene} replace the objects with pre-built CAD models, resulting in significant differences from the observed scenes. (c) In contrast, our proposed approach focuses on part-level reconstruction, replacing object parts with aligned primitive shapes. This yields interactive scenes with finer details and better alignment with the physical environment.}
    \label{fig:motivation}
\end{figure}

By enriching the reconstructed scenes with interactions that allow robots to anticipate action effects and verify their plans without executing them in the physical world, Han \etal proposed a novel task of reconstructing interactive scenes that can be imported into ROS-based simulators~\cite{han2021reconstructing,han2022scene}, which is crucial for long-horizon \ac{tamp}~\cite{kaelbling2011hierarchical,jiao2022sequential,su2023sequential}. This approach involves using a 3D panoptic mapping method to reconstruct scenes from RGB-D data, segmenting objects, and representing them as 3D meshes (\cref{fig:motive_panoptic}). The segmented object meshes are then replaced with CAD models from a database, which provide actionable information about how robots can interact with them. This approach facilitates the emulation of complex interactions in simulated environments. \cref{fig:motive_object} shows a reconstructed interactive scene with objects replaced by CAD models.

Despite the successful attempt to reconstruct interactive scenes, there are challenges in reproducing the observed scenes with adequate fidelity. As shown in \cref{fig:motive_panoptic,fig:motive_object}, noisy perception and severe occlusions in the scans often result in unsatisfactory CAD replacements. The database's limited number of CAD models further compounds this issue, as they cannot account for the wide variety of objects robots may encounter. As a result, the reconstructed interactive scenes may lack realism and fail to represent the physical scenes accurately. 

\begin{figure*}[t!]
    \centering
    \includegraphics[width=\linewidth]{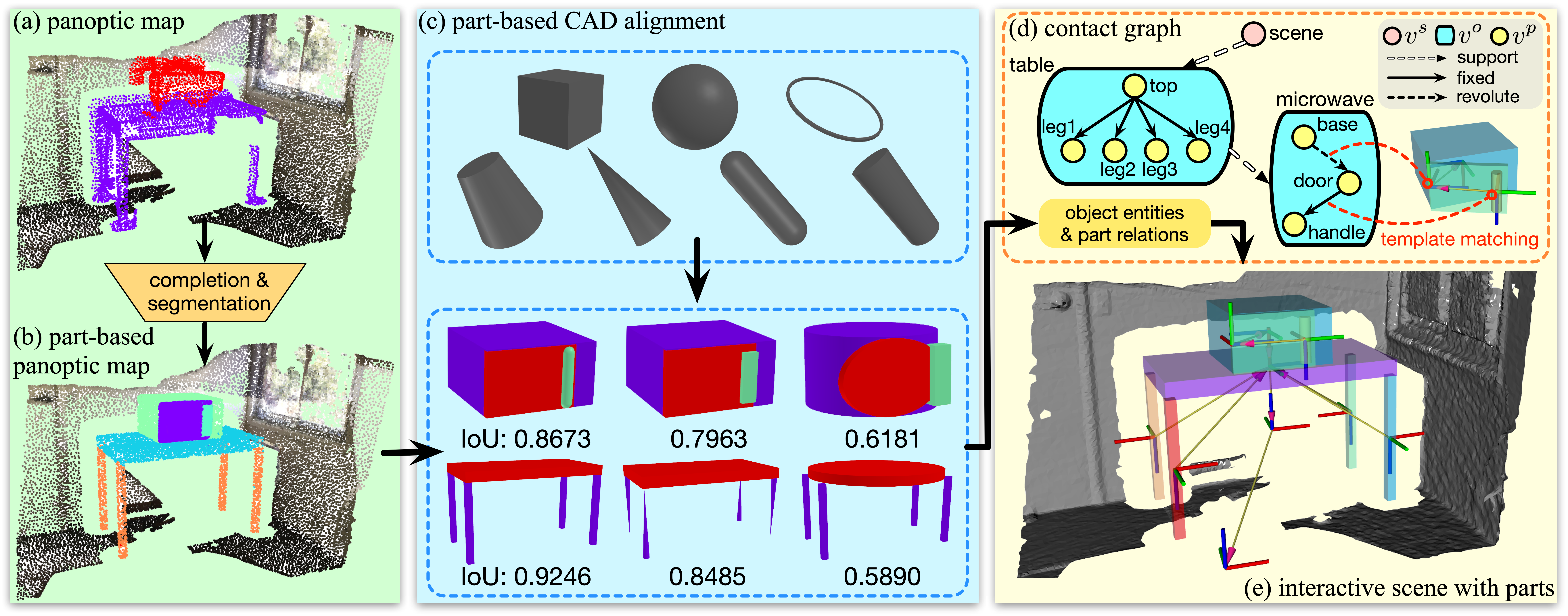}
    \caption{\textbf{System architecture for part-level interactive scene reconstruction.} (a) The initial step involves completing and segmenting the point clouds of the noisily segmented 3D objects, resulting in (b) a part-based panoptic map. (c) Each completed object part is replaced with the most aligned primitive shape. The optimal combination of part alignments, determined by the highest \acs{iou}, is selected to (d) estimate the kinematic relations among the parts. (e) The replaced object parts and their relations are compiled into a \acs{urdf} representation, capturing the kinematics of objects and the scene. This \acs{urdf} can be imported into various simulators for \ac{tamp} tasks.}
    \label{fig:pipeline}
\end{figure*}

In this work, we aim to improve the fidelity of reconstructed interactive scenes by extending the approach of Han \etal~\cite{han2021reconstructing,han2022scene}. We propose a part-level reconstruction strategy that focuses on reconstructing scenes by replacing object CAD models at the part level instead of the object level. We employ a semantic point cloud completion network to decompose and complete each noisily segmented object into parts. Next, we perform part-level CAD replacement, including aligning primitive shapes to individual parts and estimating their kinematic relations. This pipeline (see \cref{fig:pipeline}) enables the creation of a kinematics-based scene graph that captures the geometry, semantics, and kinematic constraints of the environment, facilitating more realistic robot interactions. Our part-level reconstructed interactive scenes (see \cref{fig:motive_part}) closely align with the physical scenes, providing the fidelity that can enable more accurate simulations of robot interactions.

\subsection{Related work}

Constructing an effective \textbf{scene representation} that facilitates robot mobile manipulation planning is an open problem. Traditional semantic mapping and \ac{slam} methods produce flat representations primarily suited for navigational tasks~\cite{cadena2016past}. In contrast, graph-based representations such as scene grammar~\cite{huang2018cooperative,jiang2018configurable,chen2019holistic++} and 3D scene graphs~\cite{armeni20193d,wald2020learning,rosinol20203d,han2021reconstructing} offer more structural and contextual information, enabling more versatile robot planning capabilities. In particular, Han \etal~\cite{han2021reconstructing,han2022scene} introduced a contact graph that can be automatically converted into a \ac{urdf}, providing robots with interpretable kinematic relations~\cite{jiao2021efficient,jiao2021virtual,jiao2022sequential,su2023sequential}. Building upon this work, our approach extends the field by introducing a part-level CAD replacement algorithm for reconstructing interactive scenes.

In the domain of object modeling, \textbf{part-based} approaches leverage computer vision techniques to track movements among object parts~\cite{weng2021captra,huang2021multibodysync}, exploit contextual relations from large datasets~\cite{mo2019partnet,bokhovkin2021towards,xu2022partafford} or develop data-efficient learning methods~\cite{liu2023self,liu2023semi}. These approaches aim to recognize and segment object parts, enhancing the understanding of complex object structures, but they do not yield a holistic representation of a scene that encompasses multiple objects.  

Part-level \textbf{primitive shapes}, such as spheres, cylinders, and cuboids, have been utilized to simplify the modeling of complex objects in images~\cite{zhu2005textons,wu2010learning} and scenes~\cite{miller2003automatic,lin2020using}. However, these part-level representations typically focus on the static aspect of the perceived environment, lacking the essential kinematic information required for robots to actively interact with their surroundings. This absence of kinematic information is a common limitation in the robotics community. To bridge this crucial information gap, we propose a novel part-level framework that leverages primitive shapes and estimates their kinematic relations.

\subsection{Overview}

This paper is organized as follows: \cref{sec:representation} presents our kinematics-based scene graph representation. \cref{sec:scene} introduces the part-level CAD replacement algorithm. In \cref{sec:exp}, we demonstrate the efficacy of the proposed method in various settings. Finally, \cref{sec:conclude} concludes the paper.

\section{Kinematics-based Scene Representation}\label{sec:representation}

We extend the contact graph ($cg$) introduced by Han \etal~\cite{han2021reconstructing} to represent 3D indoor scenes by incorporating scene entity parts and their kinematic information. The $cg = (pt, E)$ consists of a parse tree ($pt$) and a set of proximal relations ($E$). The parse tree organizes scene entity nodes ($V$) hierarchically based on supporting relations ($S$), while the proximal relations capture the relationships between objects. Each object node in $V$ includes attributes describing its semantics and geometry while supporting and proximal relations impose constraints to ensure physically plausible object placements.

To enhance the $cg$, we introduce an additional attribute, denoted as $pt^p$, to each object node $v \in V$. This attribute represents a per-object part-level parse tree ($pt^p$), which organizes part entities ($V^p$) along with their kinematic relations ($\mathcal{J}$). The part entities and kinematic relations are defined as follows.

The set of \textbf{part entity nodes}, denoted as $V^p = {v^p}$, represents all part entities within an object. Each part entity, $v^p = \langle l, c, M, \Pi \rangle$, encodes a unique part instance label ($l$), a part semantic label ($c$) such as ``table leg,'' a geometry model ($M$) in the form of a triangular mesh or point cloud, and a set of surface planes ($\Pi$). The surface planes are represented as $\Pi={(\boldsymbol{\pi}^k, U^k)}$, where $U^k$ is a list of 3D vertices defining a polygon that outlines the plane $\bm{\pi}^k$. The plane $\boldsymbol{\pi}^k$ is represented by a homogeneous vector $[{\boldsymbol{n}^k}^T, d^k]^T \in \mathbb{R}^4$ in projective space. The unit plane normal vector is denoted as $\boldsymbol{n}_i^k$, and the equation ${\boldsymbol{n}^k}^T \cdot \boldsymbol{u} + d^k=0$ describes the constraint satisfied by any point $\boldsymbol{u} \in \mathbb{R}^3$ on the plane.

The set of \textbf{kinematic relations}, $\mathcal{J} = {J_{p,c}}$, represents the parametric joints between part entities within an object. A joint, $J_{p,c} = \langle t_{p,c}, T_{p,c}, \mathcal{F}{p,c} \rangle$, exists between a parent part ($v_p$) and a child part ($v_c$). The joint encodes the joint type ($t{p,c}$), the parent-to-child transformation ($T_{p,c}$), and the joint axis ($\mathcal{F}_{p,c} \in \mathbb{R}^3$).

In this paper, we consider three types of joints:
\begin{itemize}[leftmargin=*,noitemsep,nolistsep,topsep=0pt]
    \item \texttt{fixed joint}: Represents a rigid connection between two parts, such as a table top and a table leg.
    \item \texttt{prismatic joint}: Indicates that one part can slide along a single axis with respect to the other, as seen in an openable drawer and a cabinet base.
    \item \texttt{revolute joint}: Represents a joint where one part can rotate around a single axis in relation to another part, like the door and base of a microwave.
\end{itemize}

Establishing a kinematic relation between two parts $v_p$ and $v_c$ requires them to be in contact with each other by satisfying the following constraints:
\begin{equation}
    \small
    \begin{aligned}
        \exists~(\boldsymbol{\pi}^i_p, U^i_p) & \in \Pi_p,~(\boldsymbol{\pi}^j_c, U^j_c) \in \Pi_c, \\
        \text{s.t.}~\text{Align}\left(\boldsymbol{\pi}_p^i, \boldsymbol{\pi}_c^j\right) & \stackrel{\text{def}}{=} \text{abs}\left({\boldsymbol{n}_p^i}^T \boldsymbol{n}_c^j \right) \geq \theta_{a},\\
        \text{Dist}\left(\boldsymbol{\pi}_p^i, \boldsymbol{\pi}_c^j\right) & \stackrel{\text{def}}{=}  \frac{1}{|U^j_c|} \sum_{\bm{u} \in U^j_c}
            {\boldsymbol{n}^i_p}^T \boldsymbol{u} + d_p^i \leq \theta_{d},\\
        \text{Cont}\left(U_p^i, U_c^j\right) & \stackrel{\text{def}}{=} \text{A}\left(U_p^i \cap \text{proj}_{p,i}(U_c^j)\right)/\text{A}\left(U_c^j\right) \geq \theta_{c},
    \end{aligned}
    \label{eqn:support_relation}
\end{equation}
where:
\begin{itemize}[leftmargin=*,noitemsep,nolistsep,topsep=0pt]
    \item $\text{Align}\left(\boldsymbol{\pi}_p^i, \boldsymbol{\pi}_c^j\right)$ defines the alignment between two surface planes, $\text{abs}(\cdot)$ computes the absolute value, and $\theta_{a}$ is the threshold to determine a good alignment ($\theta_{a}$ = 1 for a perfect alignment where two planes are parallel);
    
    \item $\text{Dist}\left(\boldsymbol{\pi}_p^i, \boldsymbol{\pi}_c^j\right)$ defines the distance between the surface planes by averaging the distances from vertices of polygon $U_c^j$ (that outlines the surface plane $\bm{\pi}_c^j$) to plane $\bm{\pi}_p^i$, $|U|$ is the number of vertices, and $\theta_{d}$ is the maximum distance allowed;
    
    \item $\text{Cont}\left(U_p^i, U_c^j\right)$ defines the contact ratio, $\text{A}(\cdot)$ computes the area of a polygon, $\theta_c$ is the minimum sufficient contact ratio, $\cap$ computes the intersection between two polygons, and $\text{proj}_{p,i}(U_c^j)$ projects polygon $U_c^j$ onto the plane $\bm{\pi}_p^i$ by projecting each vertex in $U_c^j$ onto $\bm{\pi}_p^i$:
\end{itemize}
\begin{equation}
    \hat{\bm{u}}_{c}^{j} = \bm{u}_{c}^{j} - {\bm{n}_p^i}^T(\bm{u}_{c}^{j} - \bm{u}_p^i)~\bm{n}_p^i,~\forall \bm{u}_c^j \in U_c^j,
    \label{eqn:projection}
\end{equation}
where $\hat{\bm{u}}_{c}^{j}$ is the projected point of $\bm{u}_{c}^{j}$ on the plane $\bm{\pi}_p^i$, and $\bm{u}_p^i$ is an arbitrary point on $\bm{\pi}_p^i$.

By definition, a $cg$ augmented with object parts and kinematics sufficiently defines objects' semantics, geometry, and articulations in a scene. Crucially, such a representation is also naturally compatible with the kinematic tree and could be seamlessly converted to a \ac{urdf} for various downstream applications. Leveraging $cg$, a robot can reason about the action outcomes when it interacts with (articulated) objects.

\section{Part-level CAD Replacement}\label{sec:scene}

We aim to replace the segmented and completed part entities (as shown in \cref{fig:pipeline}b) with best-aligned primitive shapes while estimating their kinematic relations, and construct a part-level contact graph $cg$ as defined in \cref{sec:representation}.

\subsection{Individual part alignment}\label{sec:ind_align}

For each individual part, we select a primitive shape with a sufficient level of similarity and calculate a 6D transformation to align the shape to the part. Given a part entity with a point cloud segment $P$, we find an optimal primitive shape $M^*$ from a finite set of primitive candidates $\mathcal{M}^c$ and an optimal 6D transformation $T_{ind}^* \in SE(3)$ that aligns $M^*$ with $P$. We obtain $\mathcal{M}^c$ based on pre-defined primitive shape templates and a 3D scaling vector estimated from the minimum-volume oriented 3D bounding box of $P$~\cite{dimitrov2009bounds}. The optimization problem can be formulated as follows:
\begin{equation}
    \small
    M^*, T_{ind}^* = \min_{M_i \in \mathcal{M}^c, T \in SE(3)} \frac{1}{|h(M_i)|} \sum_{\bm{u} \in h(M_i)} d_{P}(T_i \circ \bm{u}),
    \label{eqn:part_alignment}
\end{equation}
where $h(M_i)$ is a set of evenly sampled points on the surface of the CAD model $M_i$, $d_P(\bm{u})$ is the distance from a sampled point $\bm{u}$ to the closest point in $P$, and $T_i \circ \bm{u}$ is the position of point $\bm{u}$ after applying transformation $T_i$.

To solve this optimization problem, we compute the optimal transformation $T_i^*$ for each primitive candidate $M_i$ using the iterative closest point method~\cite{besl1992method}. Then $M^*$ is the primitive candidate with the smallest minimum total distance among all candidates $\{M_i\}$, and $T_{ind}^*$ is the corresponding optimal transformation in $\{T_i^*\}$. 

\subsection{Kinematic relation estimation}\label{sec:construct_cgpp}

After replacing part entities with primitive shapes based on individual shape alignment results, we estimate the parent-child contact relations and kinematics (\ie, parametric joints) between parts to obtain a per-object part-level parse tree $pt^p$. To initialize a part node $v^p$:
\begin{enumerate}[leftmargin=*,noitemsep,nolistsep,topsep=0pt]
    \item We acquire its part-level semantic label $c$, instance label $l$, and point cloud $P$.
    \item We replace its point cloud $P$ with a primitive shape $M$, as described in \cref{sec:ind_align}.
    \item We extract surface planes $\Pi$ from $M$ by iteratively applying RANSAC~\cite{taguchi2013point}.
\end{enumerate}

For a set of part entity nodes $V^p$ corresponding to an object, we estimate the structure of $pt^p$, \ie, the optimal parent-child contact relations among the parts of an object ${S^p}^* = \{s_{p,c}\}$ in terms of \cref{eqn:support_relation}. We formulate an optimization problem to maximize the overall contact scores $\text{Cont}\left(\cdot, \cdot\right)$ while satisfying the constraints in \cref{eqn:support_relation}:
\begin{equation}
    \begin{aligned}
        {S^p}^* &= \operatorname*{argmax}_{S^p}~\sum_{s_{p, c} \in S^p} \max_{i,j}\left(\text{Cont}\left(U_p^i, U_c^j\right)\right),\\
        \text{s.t.}~&\text{Align}\left(\boldsymbol{\pi}_p^i, \boldsymbol{\pi}_c^j\right) \geq \theta_{a},~\forall p,c, ~\exists i,j,\\
        & \text{Dist}\left(\boldsymbol{\pi}_p^i, \boldsymbol{\pi}_c^j\right) \leq \theta_{d},~\forall p,c, ~\exists i,j.\\
    \end{aligned}
    \label{eqn:pt_inference}
\end{equation}

We solve this optimization problem in two steps:
\begin{enumerate}[leftmargin=*,noitemsep,nolistsep,topsep=0pt]
    \item We construct a directed graph with nodes in $V^p$. By traversing all pairs of nodes, our algorithm adds an edge $s_{p,c}$ from node $v_p$ to node $v_c$ to the graph if they satisfy the constraints in \cref{eqn:pt_inference}, with the edge's weight set to $\max_{i,j}\left(\text{Cont}\left(U_p^i, U_c^j\right)\right)$.
    \item We find the optimal parent-child relations ${S^p}^*$. Although the constructed graph entails all possible contact relations among entities, it may not be in the form of a parse tree since the indegree of a node could be greater than 1 (\ie, a node has multiple parents), violating the definition of a rooted tree. Finding the optimal parent-child relations is equivalent to finding a directed spanning tree of maximum weight in the constructed graph, known as an arborescence problem. We adopt Edmonds' algorithm~\cite{edmonds1967optimum} to solve this problem.
\end{enumerate}

\begin{figure}[t!]
    \centering
    \includegraphics[width=\linewidth]{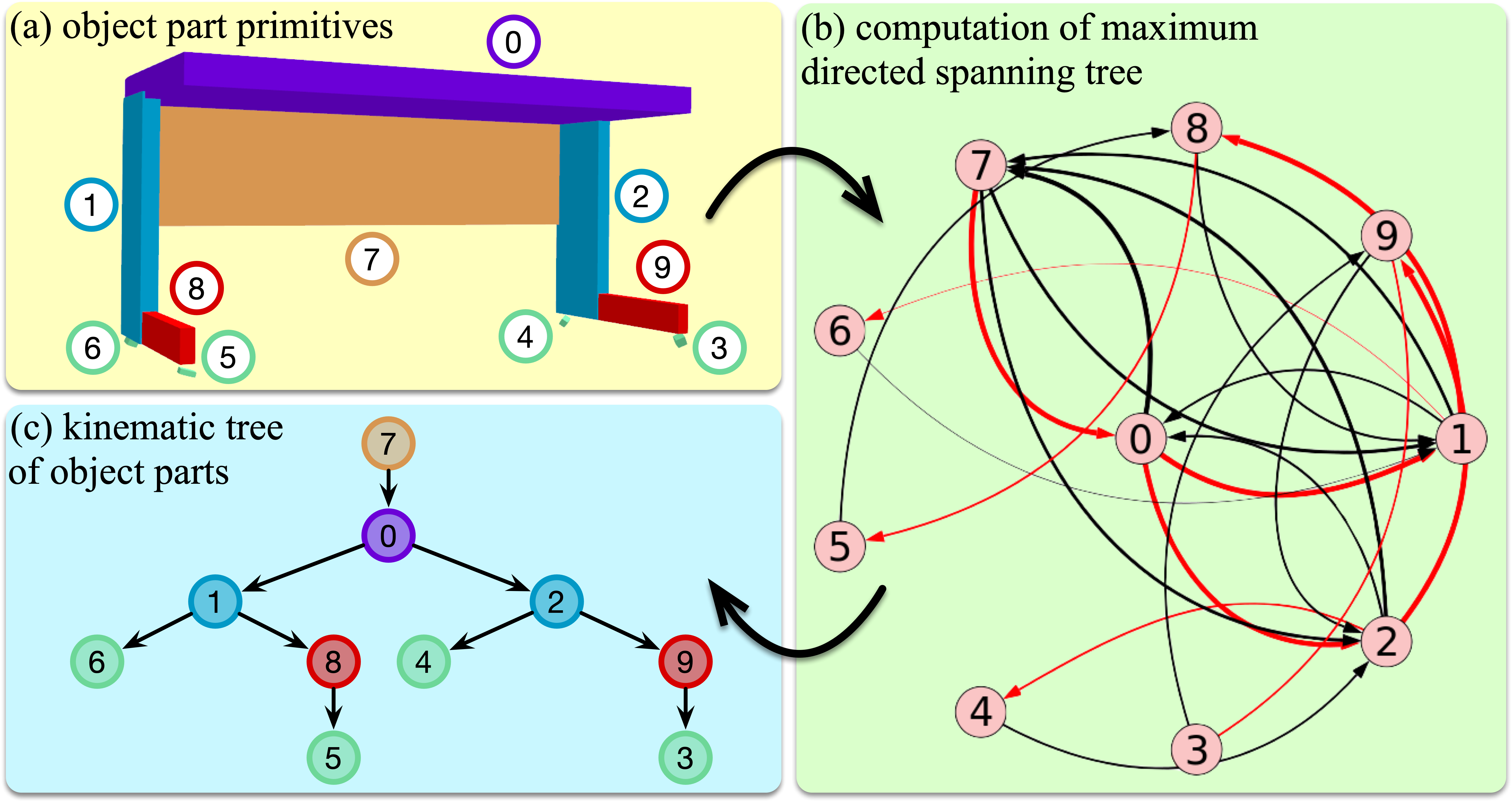}
    \caption{\textbf{Kinematic relation estimation among parts.} (a) The set of primitive shapes that best match the part entities of a table. (b) Based on the largest contact score $\text{Cont}\left(U_p^i, U_c^j\right)$ between every pair of parts (indicated by the edge's weight), the most probable connectivity between parts can be found by computing the maximum directed spanning tree, \ie, the red edges. (c) The computed kinematic relations among the parts from parent to child.}
    \label{fig:alg_fig}
\end{figure}

Next, we estimate parameterized joints $\mathcal{J}$ for all parent-child relations in ${S^p}^*$ by matching the primitive parts to a library of articulated templates. This involves determining the joint types, joint axes, and joint poses based on their semantic labels, parent-child relations, and geometries. For example, a microwave door should be connected to its base with a revolute joint, which is usually located at the rim of its base. \cref{fig:alg_fig} presents a complete example of estimating the kinematic relations among table parts.

\subsection{Spatial refinement among parts}

We can further perform a refinement process to adjust transformations in $\mathcal{J}$ so that parts forming parent-child pairs are better aligned. This step reduces penetration between parts.

The spatial refinement algorithm, detailed in \cref{alg:spatial_refinement}, performs refinements between parts in a top-down manner, given the input parse tree $pt^p$ of an object, to avoid conflicts. The function \texttt{getEdgeTransform} retrieves the relative transformation $T_{p,c}$ from the parse tree. Then, \texttt{getAlignedPlanes} pairwise compares surface planes in $v_p$ and $v_c$, and selects roughly-aligned normal vectors of planes for downstream transformation refinement. Next, the function \texttt{refineTF} refines the rotation of $v_c$ by computing a translation-free refinement transformation $T^r_c$ that aligns a set of normal vectors $X_c$ to another set $X_p$. Finally, \texttt{updateEdgeTransform} makes necessary updates to elements in $pt^p$ (\ie, $T_{p,c}$ in $\mathcal{J}$) using the refined transformation.

The optimization problem in \texttt{refineTF} is formulated as:
\begin{equation}
    \small%
    \begin{aligned}
        T^* &= \operatorname*{argmin}_{T \in SE(3)} \sum_{\bm{u}_i^p \in X_p, \bm{u}_i^c \in X_c}\left|\left|T \circ \bm{u}_i^c - \bm{u}^p_i \right|\right|_2^2, \\
        \text{s.t.}&~ T =
            \begin{bmatrix*}
                R_{3 \times 3} & \bm{0}\\
                \bm{0}^T & 1
            \end{bmatrix*},
        \label{eqn:refinement}
    \end{aligned}
\end{equation}
where $R_{3 \times 3}$ is a rotation matrix, $\bm{u}_i^p$ and $\bm{u}_i^c$ are a correspondent pair of normal vectors close to each other in direction. This optimization problem is equivalent to a point set registration problem, for which we find the optimal solution using the Kabsch algorithm~\cite{kabsch1976solution}.

Finally, we combine the part-based representation $pt^p$ for all objects into a single contact graph $cg$ of the scene. Following Han \etal~\cite{han2021reconstructing,han2022scene}, we build object entity nodes and estimate the inter-object supporting and proximal relations. Then, we refine the pose of each whole object based on the supporting relations. The resulting $cg$ effectively organizes parts of all objects in the scene with kinematic information and can be converted into a kinematic tree in \ac{urdf} format to directly support robot interactions.

\begin{algorithm}[thb!]
    \caption{Spatial refinement among parts}
    \label{alg:spatial_refinement}
    \LinesNumbered
    \SetKwInOut{KIN}{Input}
    \SetKwInOut{KOUT}{Output}
    
    \KIN{a part-level parse tree $pt^p$}
    \KOUT{$pt^p$ with refined transformations}
    
    \nosemic $q \leftarrow \texttt{Queue}()$\;
    \nonl \textcolor{blue}{// add children of root of $pt$ to queue $q$}\;
    \ForEach{$v_c \in pt^p.root.children$}{
        $q.\texttt{push}(v_c)$\;
    }
    \While{$q$ is not empty}{
        $v_c \leftarrow q.\texttt{pop}()$\;
        $v_p \leftarrow v_c.parent$\;
        \nonl \textcolor{blue}{// get transformation from $v_p$ to $v_c$}\;
        $T_{pc} \leftarrow pt^p.\texttt{getEdgeTransform}(v_p, v_c)$\;
        
        \nonl \textcolor{blue}{// find normal vectors of nearly aligned planes}\;
        $X_p, X_c \leftarrow \texttt{getAlignedPlanes}(v_p, v_c)$\;
        \nonl \textcolor{blue}{// compute the refinement transformation of $v_c$}\;
        $T_c^r \leftarrow \texttt{refineTF}(X_c, X_p)$\;
        \nonl \textcolor{blue}{// update the transformation from $v_p$ to $v_c$}\;
        $T_{pc} \leftarrow T_{pc} T_c^r$\;
        \nonl \textcolor{blue}{// update $pt^p$ with the refined transformation}\;
        $pt^p.\texttt{updateEdgeTransform}(v_p, v_c, T_{pc})$\;
        \nonl \textcolor{blue}{// add children of $v_c$ to queue $q$}\;
        \ForEach{$v_{cc} \in v_c.children$}{
            $q.\texttt{push}(v_{cc})$\;
        }
    }
    \Return $pt^p$\;
\end{algorithm}

\begin{table*}[ht!]
    \centering
    \small
    \caption{\textbf{Quantitative comparison of geometric similarity using Chamfer distance (Cdist, the lower the better) and \acs{iou} (the higher the better)} Bold values indicate the best results between object-level baseline~\cite{han2022scene} and our part-level CAD replacement using original and completed inputs, while underlined values indicate the best results using the annotated inputs.}
    \resizebox{\textwidth}{!}{%
        \setlength{\tabcolsep}{5pt}
        \begin{tabular}{ccc xxxxxxx yyyyyyyy}
            \toprule
            & CAD & input & \multicolumn{7}{x}{SceneNN seq. ID} & \multicolumn{8}{y}{ScanNet seq. ID}\\
            \cmidrule{4-18}
            & replacement & format & 011 & 030 & 061 & 078 & 086 & 096 & 223 & 0002 & 0003 & 0092 & 0157 & 0215 & 0335 & 0560 & 0640\\
            \midrule
            \multirow{5}{*}{\rotatebox[origin=c]{90}{\textbf{Cdist.}}} & & original & $\textbf{0.189}$ & $0.759$ & $0.431$ & $0.634$ & $0.588$ & $0.508$ & $0.462$ & $0.573$ & $0.776$ & $0.392$ & $0.559$ & $0.379$ & $0.604$ & $0.329$ & $0.752$\\
            & object-level & completed & $0.329$ & $0.378$ & $0.483$ & $0.413$ & $0.601$ & $0.329$ & $0.619$ & $0.580$ & $0.710$ & $0.321$ & $0.554$ & $0.256$ & $0.663$ & $0.307$ & $0.651$\\
            & & annotation & - & - & - & - & - & - & - & $0.416$ & $0.590$ & $0.282$ & $0.321$ & $0.143$ & $0.519$ & $0.322$ & $0.554$\\
            \cmidrule{2-18}
            & \multirow{2}{*}{part-level} & completed & $0.205$ & $\textbf{0.207}$ & $\textbf{0.310}$ & $\textbf{0.187}$ & $\textbf{0.210}$ & $\textbf{0.177}$ & $\textbf{0.169}$ & $\textbf{0.202}$ & $\textbf{0.163}$ & $\textbf{0.216}$ & $\textbf{0.239}$ & $\textbf{0.192}$ & $\textbf{0.174}$ & $\textbf{0.190}$ & $\textbf{0.183}$\\
            & & annotation & - & - & - & - & - & - & - & $\underline{0.101}$ & $\underline{0.119}$ & $\underline{0.092}$ & $\underline{0.087}$ & $\underline{0.076}$ & $\underline{0.086}$ & $\underline{0.098}$ & $\underline{0.089}$\\ 
            \midrule
            \midrule
            \multirow{5}{*}{\rotatebox[origin=c]{90}{\textbf{\acs{iou}}}} & & original & $0.109$ & $0.034$ & $0.063$ & $0.028$ & $0.042$ & $0.047$ & $0.021$ & $0.021$ & $0.013$ & $0.034$ & $0.028$ & $0.033$ & $0.021$ & $0.101$ & $0.012$\\
            & object-level & completed & $0.030$ & $0.034$ & $0.087$ & $0.033$ & $0.016$ & $0.052$ & $0.040$ & $0.014$ & $0.076$ & $0.128$ & $0.027$ & $0.065$ & $0.017$ & $0.057$ & $0.018$\\
            & & annotation & - & - & - & - & - & - & - & $0.056$ & $0.100$ & $0.116$ & $0.170$ & $0.196$ & $0.067$ & $0.133$ & $0.119$\\
            \cmidrule{2-18}
            & \multirow{2}{*}{part-level} & completed & $\textbf{0.125}$ & $\textbf{0.118}$ & $\textbf{0.215}$ & $\textbf{0.157}$ & $\textbf{0.156}$ & $\textbf{0.134}$ & $\textbf{0.113}$ & $\textbf{0.191}$ & $\textbf{0.224}$ & $\textbf{0.131}$ & $\textbf{0.089}$ & $\textbf{0.192}$ & $\textbf{0.179}$ & $\textbf{0.159}$ & $\textbf{0.190}$\\
            &  & annotation & - & - & - & - & - & - & - & $\underline{0.383}$ & $\underline{0.540}$ & $\underline{0.478}$ & $\underline{0.665}$ & $\underline{0.361}$ & $\underline{0.548}$ & $\underline{0.467}$ & $\underline{0.614}$\\
            \bottomrule
        \end{tabular}%
    }%
    \label{table:iou_result}
\end{table*}

\section{Experiments}\label{sec:exp}

Experiments demonstrate that our system successfully reconstructs part-level fine-grained interactive scenes from partial scans, yielding more details of the observed scenes compared with the baseline~\cite{han2022scene} that reconstructs scenes with object-level CAD replacement.

\paragraph*{Dataset augmentation}

Due to the lack of ground-truth object geometries and part segmentation in ScanNet, we augment the dataset with the information of the CAD models in PartNet~\cite{mo2019partnet} based on the annotations in Scan2CAD~\cite{avetisyan2019scan2cad}. The kinematic joints of articulated objects are further acquired from PartNet-Mobility~\cite{xiang2020sapien}. \cref{fig:dataset}a shows some examples of augmented object models in ScanNet.

\paragraph*{Implementation details}

To detect the 3D objects from the point cloud of a scanned scene, we adopt the MLCVNet~\cite{xie2020mlcvnet} as the front end of our system, which outputs a 3D bounding box for each detected object. This model was pre-trained on the ScanNet dataset following the same train/test split described in Xie \etal~\cite{xie2020mlcvnet}. After retrieving the object point cloud inside the bounding box, we used StructureNet~\cite{mo2019structurenet} to decompose the object into parts, which incorporated point cloud completion and outlier removal during the decomposition process. Of note, our system is modularized for future integration of more powerful 3D detection/completion models.


\subsection{Part-level CAD replacement from partial 3D scan}

\paragraph*{Protocols}

We evaluate our part-level CAD replacement against a baseline that replaces interactive CADs at the object level~\cite{han2022scene} based on two criteria, geometric similarity and plausibility of kinematic estimation. The evaluations were conducted using the synthetic scans from SceneNN~\cite{hua2016scenenn} and real-world scans from ScanNet~\cite{dai2017scannet}. Specifically, we picked 7 scenes in SceneNN that were used in the baseline~\cite{han2022scene} and 8 scenes from ScanNet for the evaluations. 

\paragraph*{Geometric similarity}

We use Chamfer distance and \ac{iou} metrics to quantitatively evaluate the reconstruction results. Chamfer distance measures the point-wise distance between the surface structures of the reconstructed objects and the ground-truth scans, indicating their overall geometric similarity. \ac{iou} reflects how well the reconstructed objects align with the ground-truth objects in terms of poses and sizes. Objects are normalized to a unit box for Chamfer distance computation, and replaced objects are voxelized into a $32^3$ grid for comparison with the ground-truth voxel grid for \ac{iou}.

There were three types of input scans studied in evaluations: original RGB-D scans (\textit{original}), scans with point cloud completion and part decomposition (\textit{completed}), and annotated scans in the augmented dataset that serve as the ground-truth (\textit{annotation}). Our method was not evaluated on the \textit{original} scans as they lack part-level information. Also, sequences in the SceneNN dataset cannot be augmented as \textit{annotation} scans; thus, they are not evaluated in the corresponding setup either.

\begin{figure}[t!]
    \centering
    \includegraphics[width=\linewidth]{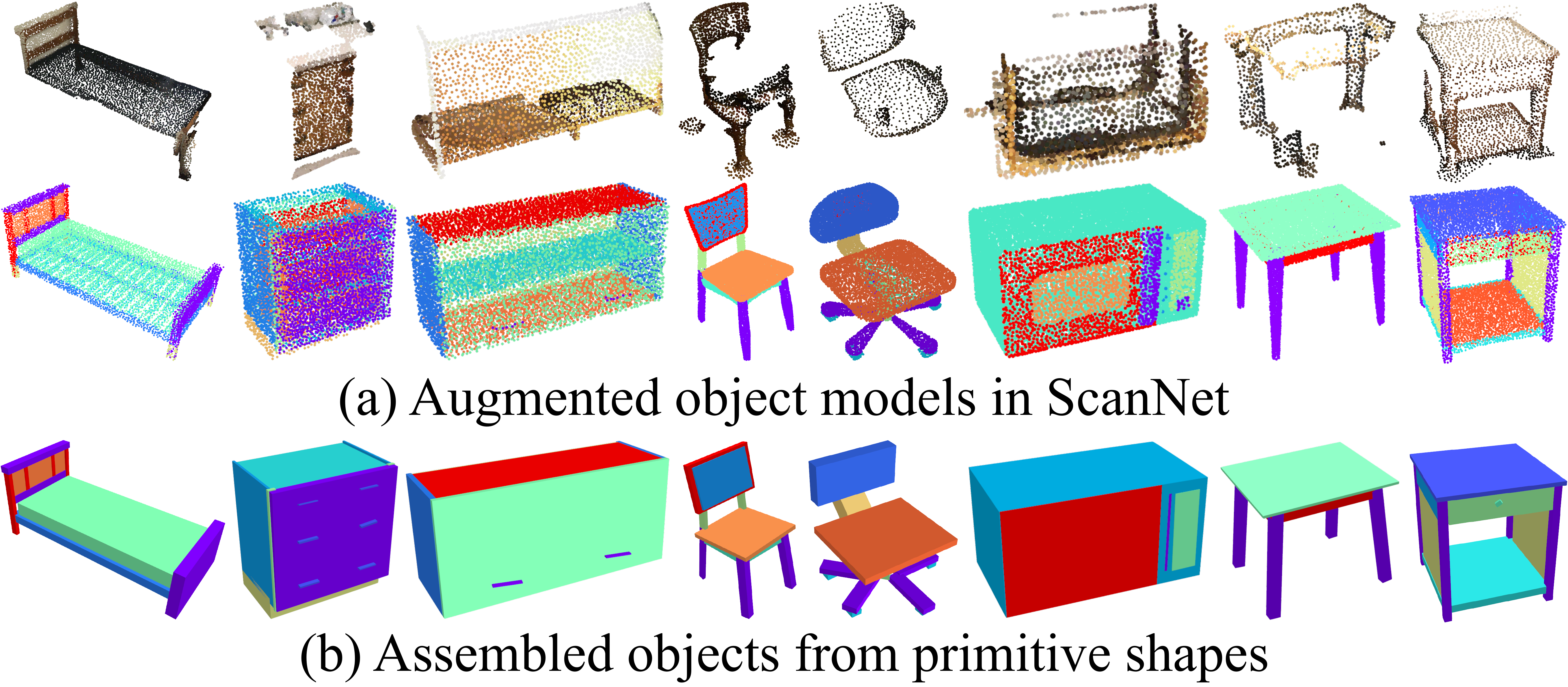}
    \caption{\textbf{Examples of augmented objects in ScanNet.} (a) Incomplete objects in ScanNet (top) are augmented by corresponding objects in PartNet with part segmentation (bottom). (b) Objects could be assembled from primitive shapes in terms of part segmentation.}
    \label{fig:dataset}
\end{figure}

The results are summarized in \cref{table:iou_result}. Looking at the \textit{original} and \textit{completed} groups, our part-based method outperforms the baseline with lower Chamfer distance and higher \ac{iou} for most sequences. This indicates the effectiveness of part-level CAD replacement for reconstructing interactive scenes and the importance of unitizing a point cloud completion model to handle noisy and incomplete scans. On the other hand, reconstructions from augmented scans with object part segmentation (see \cref{fig:dataset}b for some examples) are significantly improved (\textit{completed} \vs \textit{annotation} groups), suggesting that perception noise remains a primary challenge.

\paragraph*{Kinematic structure of object parts}

\begin{figure}[t!]
    \centering
    \includegraphics[width=\linewidth]{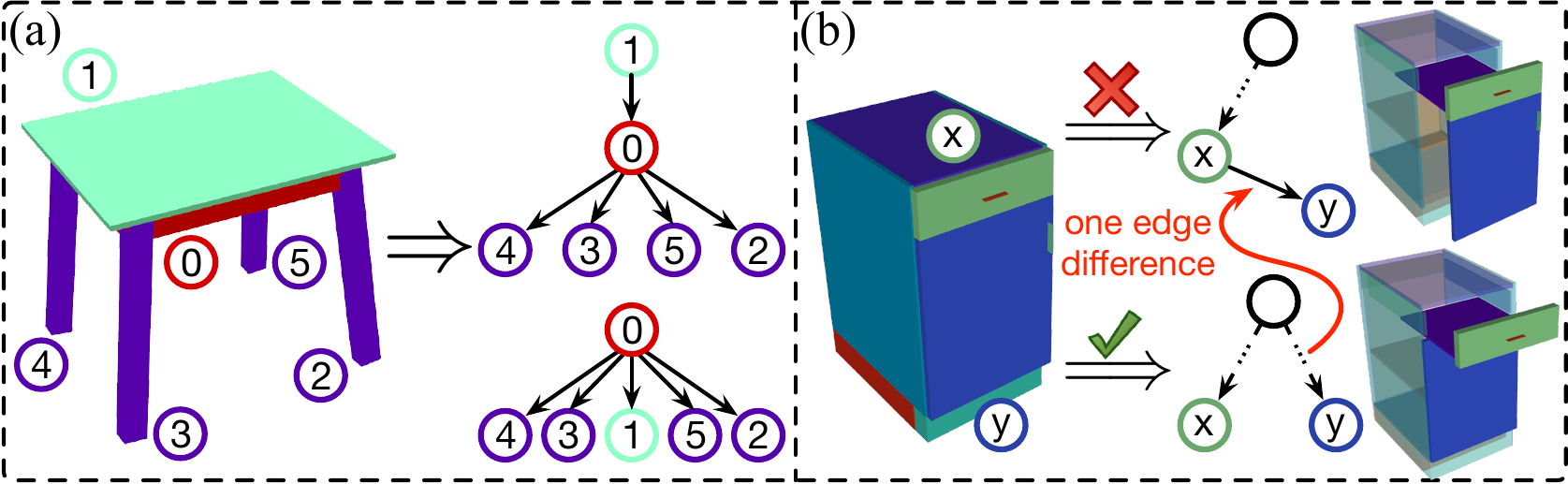}
    \caption{\textbf{Evaluation of kinematic relations.} (a) Different kinematic trees represent the parts of a table. (b) One error in kinematic structure estimation results in undesired articulation.}
    \label{fig:kinematic_eval}
\end{figure}

Evaluating the plausibility of the estimated kinematic structure of object parts is challenging due to its ambiguity. The same object can be represented by different kinematic structures (see \cref{fig:kinematic_eval}a). To address this, we manually annotate the kinematic structures of different objects. For each pair of parts in an object, we connect them with an undirected edge if we believe there is a contact relation between them. We use the \ac{map} metric to measure the alignment between the human-annotated kinematic structure and the estimated structure based on the undirected contact relations between parts (see \cref{fig:alg_fig}b). The \ac{map} metric summarizes how accurately the relations between parts (edges) are predicted.

\cref{table:kinematic} summarizes the results. Our method successfully estimates the kinematic structures of 5 object categories with high articulation, achieving \ac{map} values close to 1.0, indicating a nearly perfect match.

\begin{table}[ht!]
    \centering
    \small
    \caption{\textbf{\acs{map} of the estimated kinematic structures among object parts.}}
    \setlength{\tabcolsep}{5pt}
    \begin{tabular}{cccccc}
        \toprule
        Category & Chair & Table & Microwave & Cabinet & Bed\\
        \midrule
        \ac{map} & 0.9247 & 0.8292 & 0.9741 & 0.9592 & 0.9785\\
        \bottomrule
    \end{tabular}
    \label{table:kinematic}
\end{table}

\paragraph*{Discussions}

The results in \cref{table:iou_result} provide an indirect assessment of the kinematic transformation among parts, while \cref{table:kinematic} verifies the accuracy of estimating their parent-child relations. Although the results appear promising individually, the complex nature of kinematic relations means that even a small error can lead to significant issues. \cref{fig:kinematic_eval} showcases some typical cases, highlighting the ongoing challenge of estimating kinematic relations.

\begin{figure*}[t!]
    \centering
    \includegraphics[width=\linewidth]{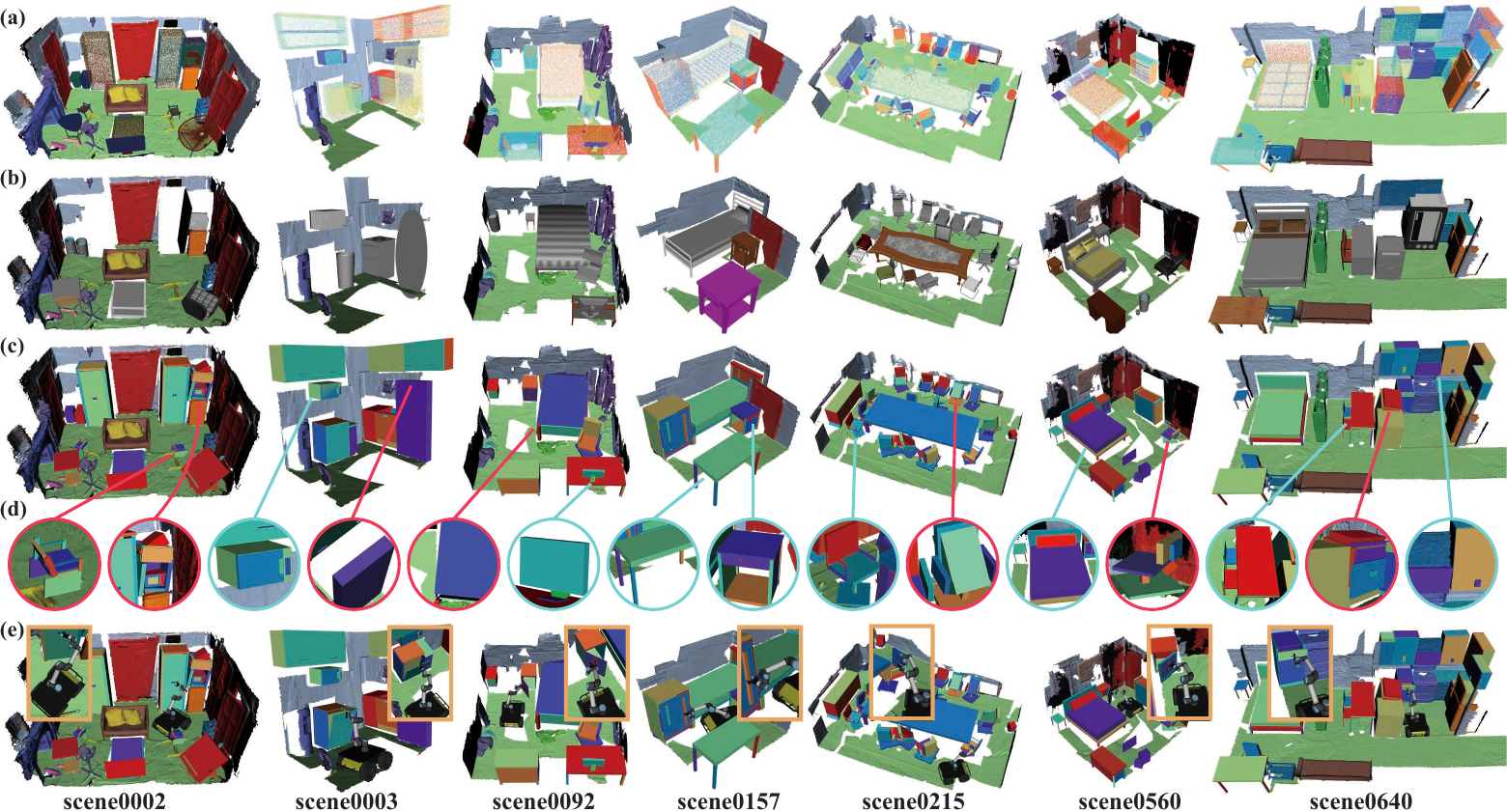}
    \caption{\textbf{Qualitative comparisons of reconstructed interactive scenes at the object level~\cite{han2022scene} and the part level (ours)}. (a) Ground-truth part-level segmentation of 7 real-world scans augmented from ScanNet~\cite{dai2017scannet}. (b) Object-level CAD replacement preserves object semantics and overall dimensions but fails to reflect geometries accurately. (c) Our part-level CAD replacement better reflects object geometries by assembling objects from primitive shapes. (d) Successful and failed part replacements/assemblies are highlighted with blue and red circles, respectively. (e) The resulting interactive scenes enable fine-grained robot interactions using \ac{tamp}).}
    \label{fig:exp3}
\end{figure*}

\subsection{Interactive scene reconstruction}

\cref{fig:exp3} provides a qualitative comparison of interactive scene reconstructions from ScanNet using object-level~\cite{han2022scene} and part-level (ours) approaches. The reconstructed scenes enable robot \ac{tamp} by leveraging the encoded kinematic relations. Our method achieves a more precise reconstruction (\cref{fig:exp3}c) compared to the baseline~\cite{han2022scene} (\cref{fig:exp3}b), as indicated by the ground-truth segmentation of the 3D scans (\cref{fig:exp3}a).

We highlight successful and failed samples in \cref{fig:exp3}d to better understand our method's performance. Failures often occur due to outliers in the part decomposition module, leading to incorrect part replacement and alignment, or when the completion module struggles with overly incomplete input point clouds (\eg, a single surface of a fridge) due to the limited information available. Many of these failure cases stem from perceptual limitations when dealing with unobserved or partially observed environments~\cite{zhu2020dark}.

Furthermore, we demonstrate that the reconstructed interactive scenes can be converted to \ac{urdf} and imported into ROS for robot-scene interactions (\cref{fig:exp3}e). The resulting contact graph containing object part geometry and kinematic relations acts as a bridge between the robot's scene perception, understanding, and \ac{tamp}~\cite{jiao2021virtual,jiao2022sequential}.

\section{Conclusion and Discussion}\label{sec:conclude}

In this work, we developed a system for reconstructing interactive scenes by replacing object point clouds with CAD models that enable robot interactions. In contrast to previous approaches focused on object-level CAD replacement~\cite{han2021reconstructing,han2022scene}, our system takes a part-level approach by decomposing objects and aligning primitive shapes to them. We achieved a more detailed and precise representation of the scene by estimating the kinematic relations between adjacent parts, including joint types (fixed, prismatic, or revolute) and parameters.

To handle noisy and partial real scans, our system incorporates a point cloud completion module to recover observed object parts before performing CAD replacement. The estimated kinematics of objects and the scene are aggregated and composed into a graph-based representation, which can be converted to a \ac{urdf}. This representation allows for reconstructing interactive scenes that closely match the actual scenes, providing a ``mental space''~\cite{zhu2020dark,fan2022artificial} for robots to engage in \ac{tamp} and anticipate action effects before execution. This capability is crucial for the success of robots in long-horizon sequential tasks.

Moreover, our system has potential applications beyond interactive scene reconstruction. It can be utilized to digitize real environments for virtual reality, creating in-situ simulations for robot planning and training~\cite{xie2019vrgym,xia2020interactive}, and facilitate the understanding of human-object interactions~\cite{li2023gendexgrasp,jiang2022chairs,liu2023reconfigurable}, among other downstream applications.

In conclusion, our part-level CAD replacement system significantly enhances the reconstruction of interactive scenes by capturing finer details and improving precision. The resulting scenes serve as a foundation for robot cognition and planning, enabling robots to navigate complex tasks successfully. Additionally, the versatility of our system opens up possibilities for various applications in virtual reality, robot planning, training, and human-object interaction studies.

\paragraph*{Limitations and future work}

Reconstructing interactive scenes, particularly at the part level, poses significant challenges that require substantial research efforts. We acknowledge the following limitations and identify potential avenues for future work.

First, real-world indoor scenes are inherently complex and are often subject to clustering, occlusions, and sensor noise. Even with scan completion methods, 2.5D RGB-D scans may still be noisy or incomplete, hindering a comprehensive understanding of the scene. Addressing this limitation requires further advancements in scan completion techniques to improve the quality of the input data.

Second, estimating an object's kinematics solely from static observations during the reconstruction process is inherently ambiguous. Existing approaches often rely on object motion cues to disambiguate kinematic relationships. However, these methods may struggle to scale effectively in larger-scale real scenes. Future research should explore novel strategies for resolving kinematic ambiguity, potentially by leveraging both static and dynamic cues or exploiting tactile information~\cite{li20233} to enhance the overall accuracy.

Third, our current system treats the interior structure of contained spaces (\eg, the space inside a cabinet) as a solid due to its unobservable nature. Human cognition excels at filling in perceptual gaps, but our system lacks this capability. Building upon our presented system, future work could integrate advanced perception and reasoning models to endow robots with similar cognitive abilities, enabling them to operate better within complex environments. Additionally, it would be valuable to develop methods that allow robots to actively probe the environment and refine reconstructed scenes, leading to more robust and detailed representations.

\textbf{Acknowledgement:}
This work is supported in part by the National Key R\&D Program of China (2021ZD0150200) and the Beijing Nova Program.

\setstretch{0.98}
\balance
\bibliographystyle{ieeetr}
\bibliography{reference_header,reference}
\end{document}